\documentclass[conference]{IEEEtran}
\IEEEoverridecommandlockouts
\usepackage{cite}
\usepackage{amsmath,amssymb,amsfonts}
\usepackage{algorithmic}
\usepackage{graphicx}
\usepackage{textcomp}
\usepackage{xcolor}
\usepackage{algorithm}         
\usepackage{algorithmic}       
\usepackage{xcolor}            
\usepackage{amsmath}           
\usepackage{booktabs} 
\usepackage{booktabs}
\usepackage{multirow}
\usepackage{array}
\def\BibTeX{{\rm B\kern-.05em{\sc i\kern-.025em b}\kern-.08em
    T\kern-.1667em\lower.7ex\hbox{E}\kern-.125emX}}
\begin{document}

\title{Multimodal Prototype Alignment for Semi-supervised Pathology Image Segmentation\\}

\author{
\IEEEauthorblockN{
Mingxi Fu\textsuperscript{1}$^{*}$,
Fanglei Fu\textsuperscript{1}$^{*}$,
Xitong Ling\textsuperscript{1}$^{*}$,
Huaitian Yuan\textsuperscript{1},
Tian Guan\textsuperscript{1}$^{\dagger}$,
Yonghong He\textsuperscript{1}$^{\dagger}$,
Lianghui Zhu\textsuperscript{1}$^{\dagger}$
}
\IEEEauthorblockA{\textsuperscript{1}Shenzhen International Graduate School, Tsinghua University\\
$^{*}$Equal contribution \quad $^{\dagger}$Corresponding authors}
}

\maketitle

\begin{abstract}
Pathological image segmentation faces numerous challenges, particularly due to ambiguous semantic boundaries and the high cost of pixel-level annotations. Although recent semi-supervised methods based on consistency regularization (e.g., UniMatch) have made notable progress, they mainly rely on perturbation-based consistency within the image modality, making it difficult to capture high-level semantic priors, especially in structurally complex pathology images. To address these limitations, we propose MPAMatch—a novel segmentation framework that performs pixel-level contrastive learning under a multimodal prototype-guided supervision paradigm. The core innovation of MPAMatch lies in the dual contrastive learning scheme between image prototypes and pixel labels, and between text prototypes and pixel labels, providing supervision at both structural and semantic levels. This coarse-to-fine supervisory strategy not only enhances the discriminative capability on unlabeled samples but also introduces the text prototype supervision into segmentation for the first time, significantly improving semantic boundary modeling. In addition, we reconstruct the classic segmentation architecture (TransUNet) by replacing its ViT backbone with a pathology-pretrained foundation model (Uni), enabling more effective extraction of pathology-relevant features. Extensive experiments on GLAS, EBHI-SEG-GLAND, EBHI-SEG-CANCER, and KPI show MPAMatch's superiority over state-of-the-art methods, validating its dual advantages in structural and semantic modeling.
\end{abstract}


\begin{figure*}[t]
\centering
\includegraphics[width=\textwidth]{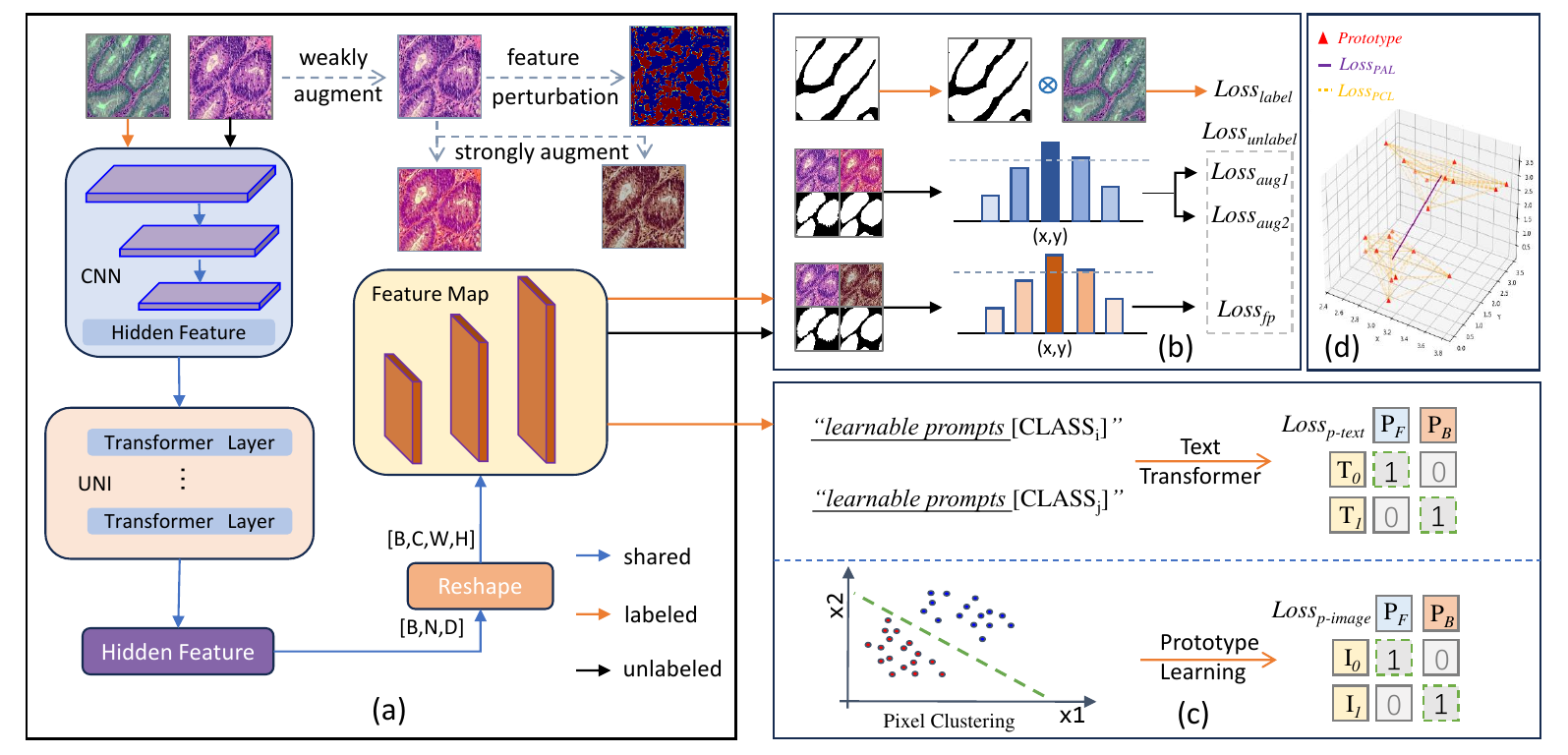}
\caption{Overall process: (a) The framework of Transunet+UNI. (b) The framework of semi-supervised segmentation method. (c) Multimodal prototype learning method. (d) The description of prototype loss calculation.}
\label{fig:model}
\end{figure*}

\section{INTRODUCTION}

With the continuous advancement of artificial intelligence in the medical field, deep learning has become a key approach for automatic analysis of whole slide images (WSIs). Among these applications, tissue semantic segmentation is one of the core tasks in computational pathology, widely used in scenarios such as disease-assisted diagnosis~\cite{frank2023accurate} and cancer subtyping~\cite{hashimoto2020multi}. Early models like FCN~\cite{long2015fully} and U-Net~\cite{ronneberger2015u} demonstrated strong performance on biomedical images, while subsequent methods such as DeepLab~\cite{chen2018encoder} further improved segmentation accuracy through dilated convolutions and multi-scale context awareness. In recent years, the introduction of Transformers has enabled better integration of local and global features such as TransUNet~\cite{chen2021transunet} and Swin-Unet~\cite{cao2022swin}. However, most existing approaches rely on fully supervised learning paradigms. In pathology, pixel-level annotation requires extensive efforts from pathologists, which greatly limits their feasibility in real clinical scenarios~\cite{campanella2019clinical}.

To address the challenge of limited annotated data in computational pathology, semi-supervised learning (SSL) has demonstrated remarkable effectiveness by leveraging both scarce labeled and abundant unlabeled samples~\cite{han2024deep}. Most state-of-the-art SSL methods adopt the consistency regularization framework, which encourages the model to produce invariant predictions under perturbations of the same input~\cite{tarvainen2017mean, hung2018adversarial, yalniz2019billion}. FixMatch~\cite{sohn2020fixmatch} innovatively combines pseudo-labeling with consistency regularization to construct pseudo-supervised objectives, while UniMatch~\cite{yang2023revisiting} further builds upon this by introducing feature-level perturbations and dual-stream augmentation strategies, achieving strong performance in semantic segmentation of natural images. However, conventional SSL segmentation methods often struggle to handle the intrinsic intra-class heterogeneity prevalent in pathological images, limiting their ability to accurately capture structural boundaries~\cite{xie2022clims}. Prototype learning has emerged as a promising solution by enhancing feature space structuring and promoting global contour awareness across whole histological slides. For instance, Yu et al.~\cite{yu2023prototypical} incorporated prototype learning into weakly supervised WSIs analysis via clustering-based prototype construction, substantially improving structural modeling. Liu et al.~\cite{liu2024pamil} further integrated prototype learning with attention mechanisms to quantify instance–prototype similarity, enabling interpretable instance-level predictions. Nevertheless, current prototype-based approaches remain largely constrained to the instance level, with limited exploration of alignment with pixel-level representations in pathology.

Nevertheless, most of these semi-supervised learning methods rely solely on visual modality information, overlooking the rich semantic information in pathological text descriptions. In recent years, pre-trained vision-language models such as CLIP~\cite{radford2021learning} and its pathology-adapted variants (e.g., PLIP~\cite{huang2023visual}, CONCH~\cite{lu2024visual}) have demonstrated great potential in incorporating textual semantic supervision. However, the text encoders in CLIP typically rely on static templates, which are insufficiently flexible to capture task-specific semantics. To enhance the task adaptability of language representations, prompt learning methods such as CoOp~\cite{zhou2022learning} have been proposed to optimize textual embeddings via learnable context prompts. These methods enable class-level semantic guidance without fine-tuning the vision-language backbone and have achieved notable success in tasks like few-shot classification.

To address the above limitations in pathological image segmentation, we propose a multimodal prototype and pixel-level label contrastive fusion framework. By constructing cross-modal semantic and visual prototypes, our method enables coarse-grained supervision to enhance global morphological structure awareness and improve segmentation performance. Specifically, we first modify the TransUNet architecture by replacing its original ViT encoder with UNI~\cite{chen2024towards}, a vision foundation model pretrained on large-scale pathology images, to better capture complex tissue morphology. In handling unlabeled data, we design a prototype-guided prediction module. On the visual side, we perform online clustering to dynamically generate structural prototypes for each class, which are fused with decoder features to refine segmentation. On the textual side, we prompt a large language model (LLM) to generate descriptive phrases for each prototype class, and extract semantic embeddings using a multimodal pathology foundation model (CONCH~\cite{lu2024visual}) to form text-guided prototypes. To enable learnability and task adaptability, we further incorporate the CoOp mechanism, allowing the text prototypes to be optimized in an end-to-end manner during training.

The main contributions of this paper are as follows: (1) We propose a multimodal prototype-guided semi-supervised segmentation framework that integrates both structural and semantic information by incorporating visual and textual prototypes, thereby improving segmentation accuracy. (2) We design a contrastive learning mechanism that combines coarse-grained (prototype-level) and fine-grained (pixel-level) supervision, and introduce a prototype-based loss function to enhance inter-class separability. (3) In the encoder-decoder architecture of the segmentation model, we replace the ViT module in TransUNet with UNI~\cite{chen2024towards}, to better capture fine-grained structures such as glandular regions and tissue boundaries in pathological images.

\section{RELATED WORK}

\subsection{Semi-supervised segmentation in histopathological image}

Semi-supervised learning (SSL) enables models to leverage abundant unlabeled data alongside limited annotations, effectively enhancing performance while reducing the dependence on costly pixel-level labels. In recent years, SSL methods have been widely applied in computer vision~\cite{shi2022semi,nguyen2020semixup,xu2022shadow,wang2021fewshot,li2021semantic}. Two common SSL strategies include consistency regularization~\cite{fan2023revisiting,zhang2022semi,lei2022semi} and self-training~\cite{ke2022three,chaitanya2023local}.

Due to the scarcity and high cost of manual annotation in computational pathology, SSL offers a practical solution for pathological images segmentation. Rashmi et al.~\cite{rashmi2024semi} proposes a hybrid strategy for tissue semantic segmentation that combines pseudo-label generation and consistency regularization. Sun et al.~\cite{sun2024semi} proposes a classification-guided semi-supervised framework that leverages multi-expert collaboration and pseudo-labeling to enhance performance in breast cancer pathological image segmentation. Han et al.~\cite{han2022multi} proposed a two-stage semi-supervised segmentation framework that first generates pseudo masks via CAM, and then trains segmentation models using multi-layer pseudo masks to effective lung adenocarcinoma segmentation. Lai et al.~\cite{lai2021semi} applied FixMatch to gray and white matter segmentation in WSIs using a two-stage pipeline: the first stage initializes the feature space learned from unlabeled images, while the second stage employs consistency regularization to enhance model generalization. However, these methods primarily rely on perturbation-based consistency within the image modality, which limits their ability to model structurally complex pathology images with ambiguous semantic boundaries.

\subsection{Multimodal Learning in histopathological image}

Pathological diagnosis is inherently a multimodal decision-making process, where clinicians rely not only on histopathological images but also on textual reports. Introducing textual modalities into image modeling enhances both interpretability and generalizability.

Recently, approaches that combine contrastive learning with text supervision have shown increasing promise in pathology image analysis. CLIP~\cite{radford2021learning} aligns images with their textual descriptions via contrastive learning and has demonstrated strong transferability in downstream tasks. Building on this, PLIP~\cite{huang2023visual}, a fine-tuned version of CLIP for pathology, aligns textual labels with WSIs to improve diagnostic region localization and data efficiency. Meanwhile, CONCH~\cite{lu2024visual} leverages over 1.17 million image-text pairs to achieve state-of-the-art results across 14 downstream tasks. Beyond image-text alignment, multimodal strategies have also been explored in modeling. Li et al.~\cite{li2024diagnostic} proposed PathTree, a framework based on a hierarchical binary class tree using expert-provided textual descriptions for hierarchical classification of complex diseases. Liu et al.~\cite{liu2024mtree} proposed an end-to-end multimodal integration of multi-scale WSI representations with clinical text, achieving outstanding performance in classification and survival prediction. Qu et al.~\cite{qu2024pathology} incorporated visual and textual priors into patch- and slide-level prompts, demonstrating excellent performance in challenging clinical tasks. Despite the strong capabilities of multimodal modeling demonstrated in the above studies, there remains a lack of effective mechanisms for leveraging fine-grained textual semantics to guide pixel-level segmentation in WSIs.

\section{METHODOLOGY}

In this section, we first introduce the overview of our whole semi-supervised segmentation framework in sections \ref{A}. The details of the proposed multimodal prototype learning are described in sections \ref{B}.

\subsection{Semi-supervised segmention framework}\label{A}

As illustrated in Fig.~\ref{fig:model}, our proposed segmentation model $S$ adopts a hybrid architecture consisting of an encoder $E$, decoder $D$, and a segmentation head.

We utilize the pathology visual foundation model (UNI~\cite{chen2024towards}) as the main encoder, initialized with weights obtained via self-supervised pretraining using DINOv2 on large-scale pathology image datasets. The model takes a $256 \times 256$ RGB image as input, which is first divided into $16 \times 16$ non-overlapping patches. Each patch is encoded into a 1024-dimensional token, resulting in a token sequence of shape $\mathbb{R}^{B \times 256 \times 1024}$, where $B$ is the batch size. These tokens serve as a global semantic representation of the image. To restore the spatial structure from the patch tokens output by the Transformer, we reshape the tokens into a 2D feature map of size $16 \times 16$, resulting in a tensor of shape $\mathbb{R}^{B \times 1024 \times 16 \times 16}$. A $3 \times 3$ convolution is then applied to reduce the channel dimension from 1024 to 512. The decoder consists of four progressive upsampling blocks with output channels configured as 256, 128, 64, and 16, respectively. Each upsampling block includes a bilinear interpolation layer followed by two $3 \times 3$ convolutional layers, gradually restoring the spatial resolution of the image. Finally, the decoder outputs a segmentation feature map of shape $\mathbb{R}^{B \times 16 \times 256 \times 256}$, which is projected to class-wise prediction logits using a $1 \times 1$ convolution, followed by a softmax activation to produce pixel-level segmentation results.

For the data with label, we adopt a hybrid loss that combines the cross-entropy loss with the Dice loss to better handle class imbalance and segmentation accuracy. The total loss $\mathcal{L}_{\text{label}}$ is defined as:
\begin{equation}
\mathcal{L}_{\text{label}} = \frac{1}{2} \left( \mathcal{L}_{\text{CE}}(\mathbf{p}, \mathbf{y}) + \mathcal{L}_{\text{Dice}}(\mathbf{p}, \mathbf{y}) \right),
\end{equation}
where $\mathcal{L}_{\text{CE}}$ denotes the standard cross-entropy loss, $\mathcal{L}_{\text{Dice}}$ denotes the Dice loss, designed to maximize the overlap between prediction and ground truth. $\mathbf{p} = \text{Softmax}(\hat{\mathbf{y}}) \in \mathbb{R}^{N \times C \times H \times W}$ is the predicted probability map, $\mathbf{y} \in \{0,1\}^{N \times C \times H \times W}$ is the one-hot encoded ground truth label.
The Dice loss is formally defined as:
\begin{equation}
\mathcal{L}_{\text{Dice}} = 1 - \frac{2 \sum_i p_i y_i + \epsilon}{\sum_i p_i + \sum_i y_i + \epsilon},
\end{equation}
where $p_i$ and $y_i$ are the flattened predicted probability and ground truth for each pixel, and $\epsilon$ is a small constant to ensure numerical stability.

For each unlabeled image $x^u$, we adopt a consistency-based strategy inspired by UniMatch~\cite{yang2023revisiting}. Specifically, the image is passed through two types of augmentations: a weak augmentation (e.g., resize) and a strong augmentation (e.g., CutMix). The weakly augmented image is used to generate a prediction $p^w$, which serves as the pseudo-label. To enhance robustness, we further apply feature-level perturbation to the extracted features of $p^w$, resulting in a perturbed prediction $p^{fp}$. In parallel, the strongly augmented view of $x^u$ is used to generate another prediction $p^s$. To enforce stronger consistency, we apply strong augmentation twice to $x^u$, yielding two perturbed views $(p^{s_1}, p^{s_2})$, both of which are aligned with the pseudo-label $p^w$ to encourage dual-stream consistency. Based on this setup, the unsupervised loss for unlabeled data is defined as:
\begin{align}
\mathcal{L}_{\text{unlabel}} &= \frac{1}{B_u} \sum_{i=1}^{B_u} \mathbf{1}(\max(p^w_i) \geq \tau) \cdot \notag \\
&\quad \left( \lambda \mathcal{H}(p^w_i, p^{fp}_i) + \frac{\mu}{2} \left[\mathcal{H}(p^w_i, p^{s_1}_i) + \mathcal{H}(p^w_i, p^{s_2}_i)\right] \right),
\end{align}
where $B_u$ is the batch size for unlabeled data, $\mathcal{H}$ denotes the cross-entropy loss, $\lambda$ and $\mu$ are weighting coefficients, and $\tau$ is the confidence threshold for pseudo-label filtering.

\subsection{Multimodal Prototype learning}\label{B}

For each semantic class $c\in\{1, \ldots, C\}$, we define $K$ prototypes $\{\mu_{c,k}\}_{k=1}^K \in \mathbb{R}^M$, resulting in a total of $CK$ prototypes. Each pixel $x_n^c$ is embedded into the feature space and assigned to the most similar prototype based on similarity. The initial prototypes are obtained via feature clustering, and subsequently updated dynamically during training using online aggregation. We define the prototype alignment loss as follows:
\begin{equation}
J = \sum_{n=1}^{\tilde{N}} \sum_{c=1}^{C} \sum_{k=1}^{K} \gamma_{n}^{c,k} \left\| x_n^c - \mu_{c,k} \right\|^2,
\end{equation}
where $\tilde{N}$ denotes the number of pixels in the current batch, and $\gamma_{n}^{c,k} \in \{0, 1\}$ indicates whether pixel $x_n^c$ is assigned to the $k$-th prototype of class $c$. During training, prototype vectors are continuously updated using the average of all assigned embeddings in the current batch.

After learning, the prototypes are fused with feature representations to guide the segmentation map generation. We first compute the similarity between each pixel and the prototypes:
\begin{equation}
z_i = \varphi(x_n, \mu_{c,k}), \quad i = 1, \ldots, N,
\end{equation}
where $\varphi(\cdot)$ denotes a similarity function (e.g., cosine similarity). Next, we utilize a cross-attention mechanism to integrate per-pixel features with the class-level prototype knowledge. Specifically, we reshape the image feature map $X \in \mathbb{R}^{H \times W \times M}$ into matrix form $\tilde{X} \in \mathbb{R}^{N \times M}$ and reshape the prototypes into a matrix $U \in \mathbb{R}^{CK \times M}$. We define $Q = \tilde{X}$, and use $U$ as both $K$ and $V$. The attention weights are then computed as:
\begin{equation}
\text{Attention}(Q, K, V) = \text{softmax} \left( \frac{Q K^T}{\sqrt{d_k}} \right) V.
\end{equation}

We concatenate the attention-enhanced features $\widetilde{X}$ with the original similarity map $S$, and feed them into a convolutional layer to obtain the probability for each pixel. The predicted probability for each pixel is then binarized to generate a segmentation mask, which serves as the foreground region indicator.

To endow each prototype with stronger semantic priors, we introduce textual modality as guidance. Specifically, for each class $c$, we construct a set of structurally descriptive texts and utilize a pretrained vision-language model to extract their embeddings as the initial semantic representations of the class prototypes. For each class $c$, we define a set of $K$ textual descriptions as $T_c = \{t_c^1, t_c^2, ..., t_c^K\}$. A pretrained text encoder (e.g., Conch) is then used to extract the embedding vector of each textual input:
\begin{equation}
e_c^k = f_{\text{text}}(t_c^k), \quad e_c^k \in \mathbb{R}^d.
\end{equation}

To improve adaptability, we further introduce $L$ learnable tokens as supplementary representations for each textual prototype, and obtain the final hybrid representation for class $c$ as:
\begin{equation}
h_c^k = \text{Mean} \left( [e_c^k, p_c^{(k,1)}, \ldots, p_c^{(k,L)}] \right),
\end{equation}
where $p_c^{(k,l)} \in \mathbb{R}^d$ denote the $l$-th cooperative token of the $k$-th textual prototype for class $c$. All tokens are initialized with random parameters and jointly optimized during training. $L$ denotes the number of learnable tokens introduced into each textual prototype. For each class $c$, we concatenate all $K$ fused prototype embeddings to obtain the textual guidance prototype tensor:
\begin{equation}
\mu_c^{\text{text}} = [h_c^1, h_c^2, ..., h_c^K] \in \mathbb{R}^{K \times d},
\end{equation}
all class-wise prototype tensors are then stacked to form the complete textual prototype tensor:$\mu^{\text{text}} \in \mathbb{R}^{C \times K \times d}$. Finally, the resulting textual feature tensor is broadcasted to the above-mentioned image modality processing pipeline to compute the logits from textual prototypes.

Based on the above, we define a Prototype Alignment Loss (PAL), which encourages each pixel to be assigned to its most compatible prototype. We then compute the visual prototype loss and textual prototype loss separately:
\begin{equation}
\mathcal{L}_{\text{PAL}} = - \frac{1}{N} \sum_{i=1}^N \log \frac{\exp(z_{i,y_i})}{\sum_{j=1}^P \exp(z_{i,j})},
\end{equation}
where $N$ is the number of valid pixels, $P$ is the total number of prototypes, $y_i \in \{0, \ldots, P-1\}$ is the prototype index assigned to pixel $i$, and $z_{i,j}$ is the similarity score between pixel $i$ and prototype $j$. We also define a Prototype Contrast Loss (PCL) to enhance the separability of prototypes across different classes. This loss encourages each pixel to be closer to the prototype of its own class while being pushed away from prototypes of other classes:
\begin{equation}
\mathcal{L}_{\text{PCL}} = - \frac{1}{N} \sum_{i=1}^N \log \frac{\exp(z_{i,y_i})}{\exp(z_{i,y_i}) + \sum_{j \in \mathcal{N}_i} \exp(z_{i,j})},
\end{equation}
where $\mathcal{N}_i = \{ j \in [0, P) \mid \lfloor \tfrac{j}{K} \rfloor \neq \lfloor \tfrac{y_i}{K} \rfloor \}$ is the set of prototypes not belonging to the same class as $y_i$.
We define the total prototype loss as a weighted combination of $\mathcal{L}_{\text{PAL}}$ and $\mathcal{L}_{\text{PCL}}$:
\begin{equation}
\mathcal{L}_{\text{proto}} = \alpha_1 \cdot \mathcal{L}_{\text{PAL}} + \alpha_2 \cdot \mathcal{L}_{\text{PCL}},
\end{equation}
where $\alpha_1$ and $\alpha_2$ are the weights for the PAL and PCL losses, respectively.
The final loss function integrates supervised segmentation loss $\mathcal{L}_{\text{label}}$, unsupervised loss $\mathcal{L}_{unlabel}$, and prototype consistency:
\begin{equation}
\mathcal{L}_{\text{total}} =\alpha \cdot \mathcal{L}_{\text{proto}} + \beta \cdot \mathcal{L}_{\text{label}} + \gamma \cdot \mathcal{L}_{unlabel},
\end{equation}
where $\alpha$, $\beta$, $\gamma$ are the weights for the $\mathcal{L}_{\text{proto}}$, $\mathcal{L}_{\text{label}}$ and $\mathcal{L}_{unlabel}$, respectively.

\section{EXPERIMENT}
\subsection{Datasets}\label{AA}
We evaluate our method on four publicly available datasets commonly used in pathological image segmentation:\\
\textbf{The Gland Segmentation (GlaS) challenge dataset}
consists of 165 pathological images extracted from 16 H\&E-stained WSIs of colorectal adenocarcinoma (T3 or T4 stage). The image sizes range from $567 \times 430$ to $775 \times 522$. Each glandular region was annotated by an expert pathologist. The dataset is split into 148 training images and 17 testing images.\\
\textbf{The EBHI-SEG dataset}
is a pathological segmentation dataset of small intestinal biopsy samples. It contains 4456 image patches of size $224 \times 224$. From this dataset, we derive two subsets: \textbf{EBHI-SEG-GLAND} with 1399 patches and \textbf{EBHI-SEG-CANCER} with 795 patches, annotated by two pathologists from the Tumor Hospital of China Medical University. The images are split into 1258 training and 141 testing samples for gland, and 715 training and 80 testing samples for cancer.\\
\textbf{The MICCAI 2024 Kidney Pathology Image(KPI) segmentation challenge dataset}
is a pathological dataset derived from renal biopsy samples. It includes 861 image patches, with structures annotated by professional pathologists. The dataset is divided into 774 training images and 87 testing images.

\subsection{Implementation Details}
For all four datasets, we use Stochastic Gradient Descent (SGD) as the optimizer, with an initial learning rate of 0.01, a weight decay of 0.0001, and a momentum of 0.9. Each model is trained for 40 epochs with a batch size of 1. During training, we apply random color jittering and CutMix augmentations. All images are resized to $256 \times 256$ before being fed into the network. We follow a 9:1 split for training and testing. Within the training set, we further divide labeled and unlabeled samples at a 7:2 ratio. All experiments are conducted on a workstation equipped with NVIDIA RTX A100 GPUs. We evaluate segmentation performance using three standard metrics: mean Intersection over Union (mIoU), mean Dice coefficient (mDice), and mean class pixel accuracy (mCPA). mIoU measures the overlap between predicted and ground truth regions, mDice evaluates set similarity, and mCPA assesses pixel-wise accuracy per class. Higher values of these metrics indicate better segmentation performance. In our experiments, we set the weighting coefficients as $\alpha = 0.25$, $\beta = 0.5$, and $\gamma = 0.25$, respectively.

\subsection{Comparison with State-of-the-Art Methods}

In this section, we present the experimental results of our newly developed MPAMatch framework on four datasets: GlaS, EBHI-SEG-GLAND, EBHI-SEG-CANCER, and KPI. As our proposed segmentation model is a plug-in framework, we select several widely used segmentation backbones in medical imaging, including DeepLabv3+~\cite{chen2018encoder}, UNet~\cite{ronneberger2015u}, UNet3+~\cite{huang2020unet}, TransUnet~\cite{chen2021transunet}, and TransUNi (Transunet+Uni encoder). We also compare our method with two widely used semi-supervised learning methods: FixMatch~\cite{sohn2020fixmatch} and UniMatch~\cite{yang2023revisiting}.

As shown in Table~\ref{tab:sota}, our MPAMatch framework consistently outperforms other methods. It achieves the mDICE of 92.44\% on GlaS, 94.73\% on EBHI-SEG-GLAND, 69.05\% on EBHI-SEG-CANCER, and 91.64\% on KPI. Notably, our method outperforms previous approaches across all datasets, demonstrating superior adaptability to different pathological segmentation tasks. This strong generalization capability highlights the framework's effectiveness in dealing with domain variation and limited supervision. The results also reveal the benefit of using stronger backbones such as TransUNi, which are built upon the powerful UNI pretraining. Moreover, under the semi-supervised setting, our MPAMatch framework achieves better performance than existing approaches built on FixMatch and UniMatch, validating the effectiveness of our prototype-driven alignment strategy.

\begin{table*}[t]
\centering
\footnotesize
\caption{Performance of MPAMatch on GlaS, EBHI-SEG-GLAND, EBHI-SEG-CANCER, and KPI datasets.}
\label{tab:sota}
\begin{tabular}{
    >{\centering\arraybackslash}m{2.8cm}  
    ccc ccc ccc ccc
}
\toprule
\multirow{2.6}{*}{\textbf{Method}} & 
\multicolumn{3}{c}{\textbf{GlaS}} & 
\multicolumn{3}{c}{\textbf{EBHI-SEG-GLAND}} & 
\multicolumn{3}{c}{\textbf{EBHI-SEG-CANCER}} & 
\multicolumn{3}{c}{\textbf{KPI}} \\
\cmidrule(lr){2-4} \cmidrule(lr){5-7} \cmidrule(lr){8-10} \cmidrule(lr){11-13}
 & mDICE & mIOU & mCPA & mDICE & mIOU & mCPA & mDICE & mIOU & mCPA & mDICE & mIOU & mCPA \\
\midrule
FixMatch+Deeplabv3+   & 60.30 & 46.75 & 65.38 & 74.50 & 60.67 & 78.47 & 63.10 & 51.69 & 70.06 & 78.36 & 69.98 & 73.07 \\
FixMatch+UNet      & 70.05 & 56.90 & 72.18 & 64.61 & 52.13 & 67.92 & 58.94 & 48.00 & 68.37 & 89.97 & 83.24 & \underline{92.17} \\
FixMatch+Unet3+    & 80.83 & 69.08 & 81.81 & 71.21 & 61.72 & 73.83 & 56.56 & 46.54 & 65.72 & 48.97 & 48.00 & 50.00 \\
FixMatch+Transunet & 87.26 & 77.77 & 87.53 & \underline{94.55} & \underline{90.01} & \underline{95.02} & 68.16 & 58.98 & 75.41 & \underline{90.99} & \underline{84.80} & 90.70 \\
FixMatch+Transuni  & 90.33 & 82.69 & 91.09 & 91.60 & 85.19 & 92.19 & 66.80 & 56.78 & 74.43 & 86.01 & 78.44 & 81.90 \\
UniMatch+Deeplabv3+   & 71.57 & 57.41 & 74.65 & 91.64 & 85.41 & 92.45 & 64.55 & 53.94 & 71.51 & 89.79 & 83.06 & 88.37 \\
UniMatch+UNet      & 83.35 & 72.23 & 83.96 & 94.33 & 89.73 & 94.82 & 64.51 & 53.36 & 71.44 & 86.36 & 78.60 & 84.78 \\
UniMatch+Unet3+    & 80.37 & 68.35 & 81.66 & 92.60 & 86.76 & 93.15 & 65.63 & 55.64 & 74.29 & 83.83 & 76.07 & 79.07 \\
UniMatch+Transunet & 85.80 & 75.69 & 86.68 & 94.45 & 89.83 & 94.88 & 66.20 & 56.37 & 72.97 & 89.36 & 82.51 & 89.25 \\
UniMatch+Transuni  & \underline{91.22} & \underline{84.06} & \underline{91.52} & 93.80 & 88.69 & 94.34 & \underline{68.76} & \underline{59.32} & \underline{79.32} & 89.94 & 83.24 & 89.45 \\
\textbf{MPAMatch}  & \textbf{92.44} & \textbf{86.09} & \textbf{92.75} & \textbf{94.73} & \textbf{90.37} & \textbf{95.15} & \textbf{69.05} & \textbf{59.63} & \textbf{79.34} & \textbf{91.64} & \textbf{85.70} & \textbf{92.78} \\
\bottomrule
\end{tabular}
\end{table*}

\subsection{Interpretability Analysis}

\begin{figure*}[t]
\centering
\includegraphics[width=0.7\textwidth]{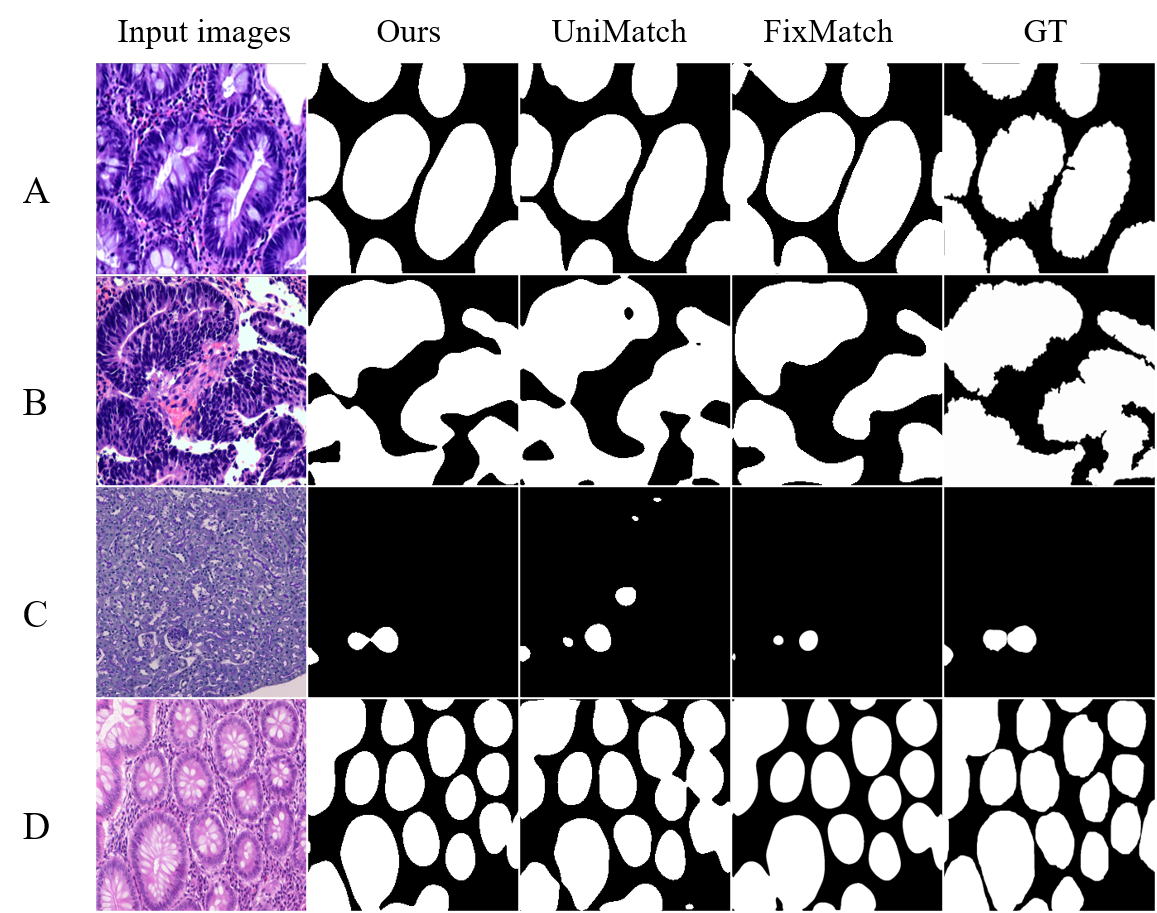}
\caption{Qualitative comparison of segmentation masks generated by FixMatch, UniMatch, and our proposed MPAMatch framework. A, B, C, and D represent the EBHI-SEG-GLAND, EBHI-SEG-CANCER, KPI and GLAS datasets, respectively.}
\label{fig:visual}
\end{figure*}

We conduct an interpretability analysis of our proposed MPAMatch semi-supervised framework. In Fig.~\ref{fig:visual}, by comparing the segmentation masks generated by different semi-supervised baselines combined with the TransUNi segmentation model, we observe that our MPAMatch framework achieves superior pathological images segmentation performance.

Specifically, in the gland segmentation task, MPAMatch is able to more accurately delineate glandular boundaries and effectively separate adjacent gland instances. In the cancer region segmentation task, MPAMatch exhibits stronger localization capability, successfully identifying malignant regions with minimal omission.

\subsection{Ablation Study}

\textbf{Effectiveness of Different Text Prompts.}
To investigate the impact of textual prompts on the MPAMatch framework, we conducted ablation experiments using various types of text descriptions. Specifically, we compared four types of prompts: (1) pathology-specific texts with varying lengths and differing semantic content across categories (\textit{P-nonsim}), (2) pathology-specific texts with consistent lengths but differing descriptions (\textit{P-simL}), (3) pathology-specific texts with both consistent lengths and consistent descriptions (\textit{P-simLD}), and (4) general (non-medical) texts with varying lengths and differing semantics (\textit{T-nonsim}).

Our results in Table~\ref{tab:text_prompt_ablation} show that pathology-specific texts with diverse lengths and semantic content (\textit{P-nonsim}) consistently achieve superior performance in segmentation tasks. This suggests that semantically rich and diverse prompts provide more informative guidance for the model. Furthermore, all pathology-specific prompts outperform general textual prompts, highlighting a strong correlation between the quality of prompt design and MPAMatch’s overall performance. These findings indicate that task-specific, domain-relevant descriptions can enhance the effectiveness of text-guided MPAMatch model in WSIs analysis.
\begin{table}[htbp]
\centering
\caption{DICE results of MPAMatch with different text prompts on four datasets}
\label{tab:text_prompt_ablation}
\begin{tabular}{ccccc} 
\toprule
\textbf{Text prompt} & \textbf{GLAS} & \textbf{EBHI-G} & \textbf{EBHI-C} & \textbf{KPI} \\
\midrule
\textit{P-nonsim}   & \textbf{92.44} & \textbf{94.73} & \textbf{69.05} & \textbf{91.64} \\
\textit{P-simL}     & 91.90 & 93.92 & 68.77 & 91.55 \\
\textit{P-simLD}    & \underline{92.38} & \underline{93.94} & 68.42 & 91.42 \\
\textit{T-nonsim}   & 92.23 & 93.89 & \underline{68.84} & \underline{91.61} \\
\bottomrule
\end{tabular}
\end{table}

\begin{figure}[t] 
\centering
\includegraphics[width=\linewidth]{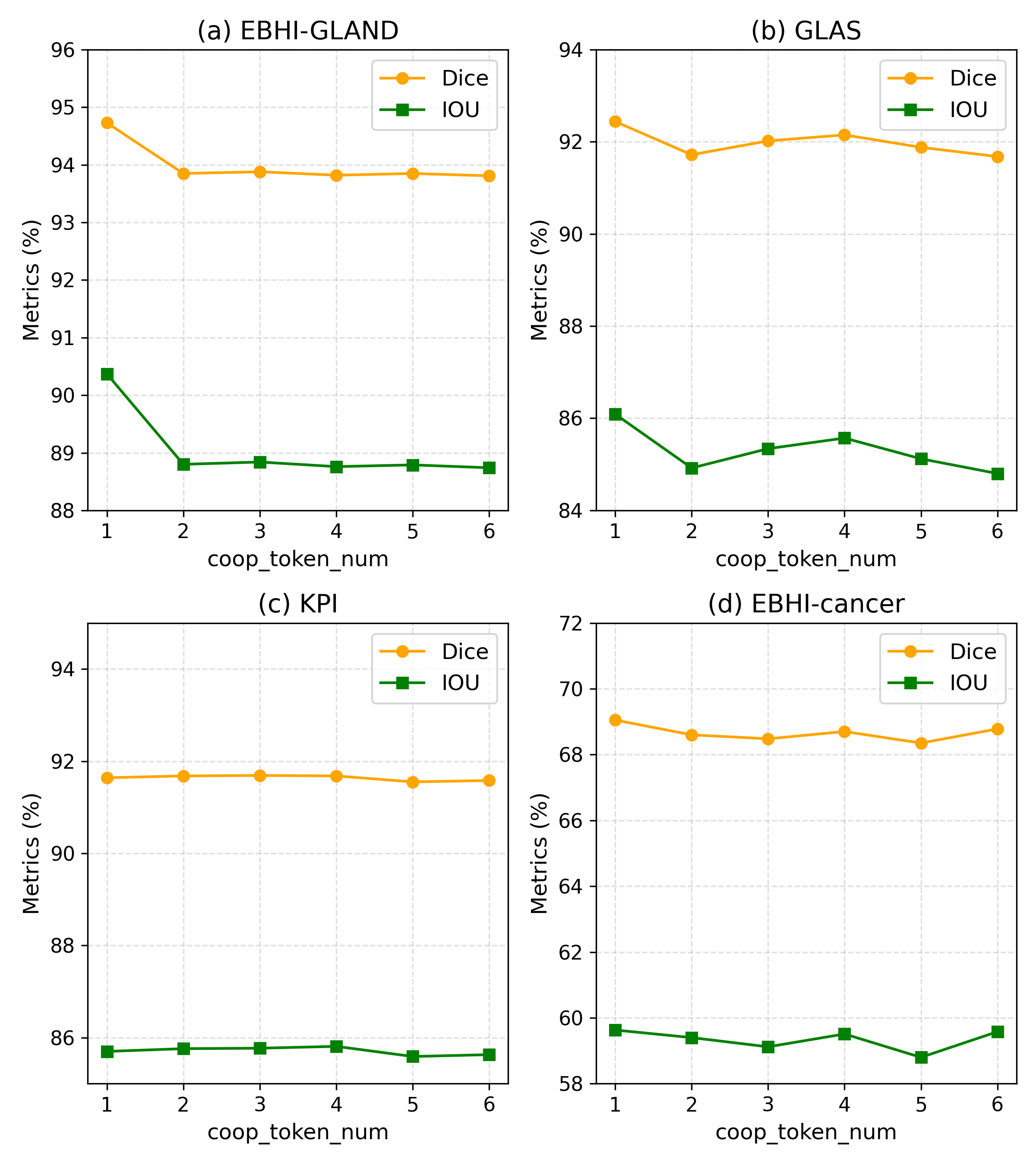} 
\caption{Influence of different numbers of coop token.}
\label{fig:iou}
\end{figure}

\textbf{Effectiveness of Different Coop Token Numbers.}
We introduce the Coop mechanism into the MPAMatch by concatenating multiple trainable tokens with image features to achieve cross-modal information fusion and fine-grained semantic guidance. To further investigate the impact of the number of Coop tokens on model performance, we conducted systematic ablation experiments under different settings of \texttt{coop\_token\_num} $\in \{1, 2, 3, 4, 5, 6\}$. During training, the trainable text tokens were concatenated with image features along the token dimension for each class and subsequently averaged to control the granularity of the fused information. In Fig.~\ref{fig:iou}, experimental results show that the model achieves the best overall performance when coop\_token\_num is set to 1, suggesting that using too many tokens may introduce redundancy or noise, thereby impairing the quality of feature aggregation. This also validates that class-level tokens are the most effective guidance signals for downstream tasks in the Coop setting.

\begin{figure}[t] 
\centering
\includegraphics[width=\linewidth]{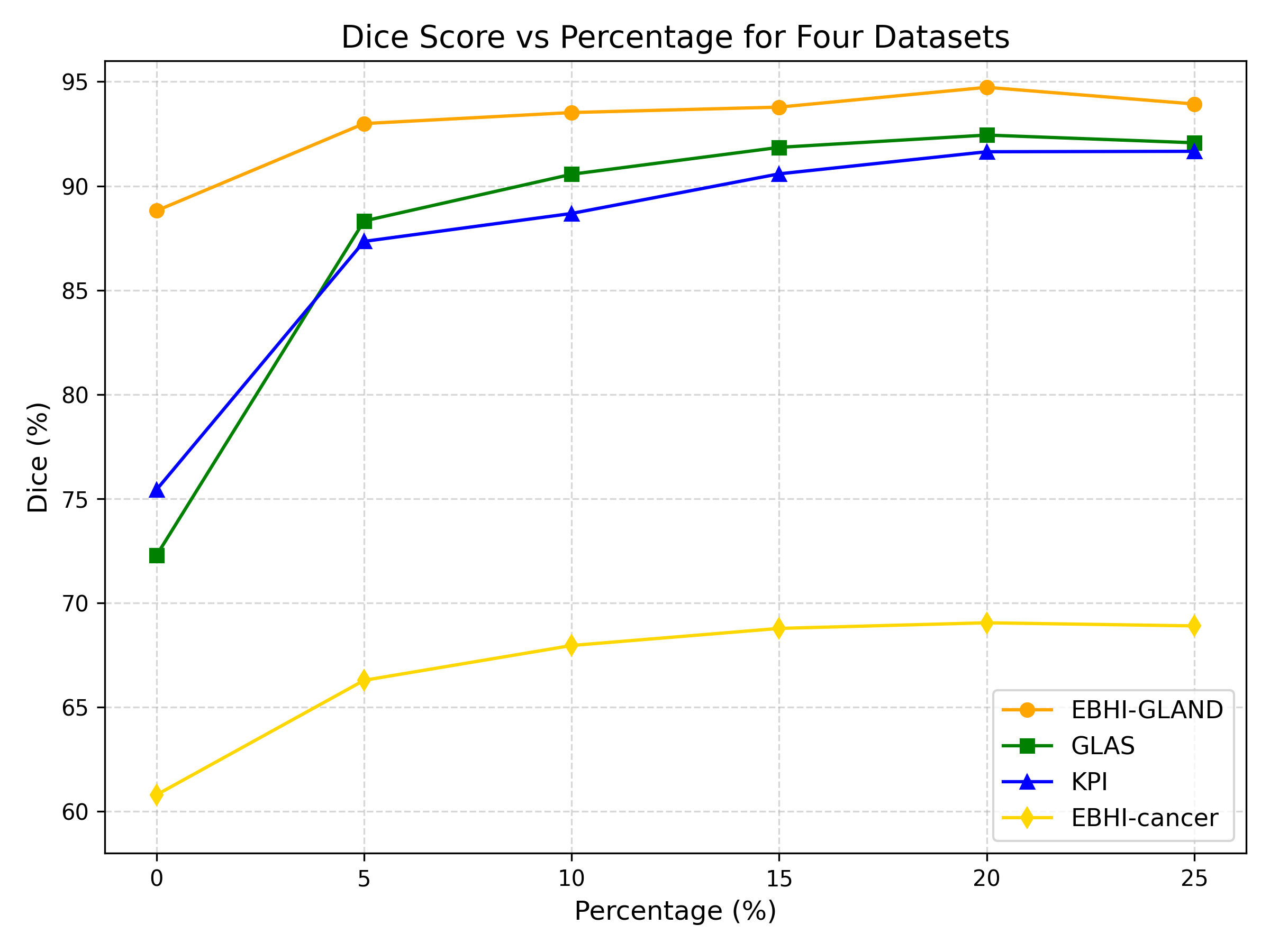} 
\caption{Influence of different percentages of unlabeled images in DICE.}
\label{fig:diceper}
\end{figure}

\textbf{Effectiveness of Different Percentages of Unlabeled Images.}
In our proposed semi-supervised framework, both labeled and unlabeled images are jointly utilized during training. To investigate how the proportion of unlabeled data affects model performance, we conducted experiments by varying the percentage of unlabeled images while keeping the total number of training images fixed. In Fig.~\ref{fig:diceper}, the results show that as the proportion of unlabeled data increases, the model's performance gradually improves. When the unlabeled data ratio reaches 15\%, the performance begins to stabilize, and in most datasets, the model achieves peak performance when the unlabeled proportion reaches 20\%. These findings demonstrate that our semi-supervised framework can achieve superior performance with a moderate amount of unlabeled data, even outperforming fully-supervised models trained with the same total number of samples.

\noindent\textbf{Effectiveness of Different thresholds.}
To investigate the impact of pseudo label quality on semi-supervised segmentation, we conduct an ablation study by varying the confidence threshold used for selecting pseudo labels from unlabeled data. Specifically, we evaluate three thresholds (90, 95, and 99) across four datasets (GLAS, EBHI-G, EBHI-C, and KPI). As shown in Table~\ref{tab:thresholds}, a moderate threshold (95) consistently achieves the best or comparable performance in terms of MDICE, MIOU, and MCPA. A lower threshold (90) introduces more noisy pseudo labels, leading to slight performance degradation due to the inclusion of less reliable predictions. Conversely, a very high threshold (99) filters out a large number of pseudo labels, resulting in insufficient supervision signals from the unlabeled data and hindering performance. These results highlight the importance of balancing pseudo label quantity and quality, and demonstrate that a threshold of 95 provides a good trade-off between label precision and coverage.
\begin{table}[htbp]
\centering
\caption{Performance comparison across different thresholds on four datasets.}
\label{tab:thresholds}
\begin{tabular}{ccccc}
\toprule
\textbf{Dataset} & \textbf{Threshold} & \textbf{MDICE (\%)} & \textbf{MIOU (\%)} & \textbf{MCPA (\%)} \\
\midrule
\multirow{3}{*}{GLAS}   & 90 & 91.91 & 85.17 & \underline{92.35} \\
                        & 95 & \textbf{92.44} & \textbf{86.09} & \textbf{92.75} \\
                        & 99 & \underline{91.94} & \underline{85.22} & 92.34 \\
\hline
\multirow{3}{*}{EBHI-G} & 90 & \underline{93.82} & \underline{88.76} & \underline{94.39} \\
                        & 95 & \textbf{94.73} & \textbf{90.37} & \textbf{95.15} \\
                        & 99 & 93.82 & 88.75 & 94.34 \\
\hline
\multirow{3}{*}{EBHI-C} & 90 & \underline{68.79} & \underline{59.50} & 77.25 \\
                        & 95 & \textbf{69.05} & \textbf{59.63} & \textbf{78.34} \\
                        & 99 & 68.36 & 58.72 & \underline{77.32} \\
\hline
\multirow{3}{*}{KPI}    & 90 & \textbf{91.66} & \textbf{85.72} & \textbf{91.15} \\
                        & 95 & \underline{91.64} & \underline{85.70} & \underline{90.78} \\
                        & 99 & 90.95 & 84.74 & 89.71 \\
\bottomrule
\end{tabular}
\end{table}

\section*{Conclusion}

This paper proposes a multimodal prototype alignment semi-supervised learning framework for pathological image segmentation, aiming to reduce the annotation burden for clinical pathologists. In our framework, we introduce both textual and visual prototypes to align with pixel-level labels, providing coarse-to-fine supervision at both structural and semantic levels. The incorporation of textual information significantly enhances the modeling of semantic boundaries. Furthermore, we reconstruct the classic segmentation architecture TransUNet by replacing its ViT backbone with a pathology-pretrained foundation model, enabling more effective extraction of pathology-relevant semantic features.

Extensive experiments conducted on the GLAS, EBHI-SEG-GLAND, EBHI-SEG-CANCER, and KPI datasets demonstrate that our method consistently outperforms several state-of-the-art semi-supervised segmentation frameworks, highlighting its effectiveness and robustness in the context of computational pathology.

\bibliographystyle{IEEEtran}
\bibliography{refs}
\end{document}